# Integrating Planning and Execution in Stochastic Domains

**Richard Dearden**
Department of Computer Science,
University of British Columbia
Vancouver, BC, CANADA, V6T 1Z4
dearden@cs.ubc.ca

**Craig Boutilier**
Department of Computer Science,
University of British Columbia
Vancouver, BC, CANADA, V6T 1Z4
cebly@cs.ubc.ca

## Abstract

We investigate planning in time-critical domains represented as Markov Decision Processes, showing that search based techniques can be a very powerful method for finding close to optimal plans. To reduce the computational cost of planning in these domains, we execute actions as we construct the plan, and sacrifice optimality by searching to a fixed depth and using a heuristic function to estimate the value of states. Although this paper concentrates on the search algorithm, we also discuss ways of constructing heuristic functions suitable for this approach. Our results show that by interleaving search and execution, close to optimal policies can be found without the computational requirements of other approaches.

## 1 Introduction

An optimal solution to a decision-theoretic planning problem requires the formulation of a sequence of actions that maximizes the expected value of the sequence of world states through which the planning agent progresses by executing that plan. Dean et al. (1993a; 1993b) have suggested that many such problems can be represented as Markov decision processes (MDPs). This allows the use of dynamic programming techniques such as *value* or *policy iteration* (Howard 1971) to compute optimal policies or courses of action. Indeed, such policies solve the more general problem of determining the best action for *every* state. Unfortunately, this optimality and generality comes at great computational expense.

Dean et al. (1993a; 1993b) have proposed a planning method that relaxes these requirements. An *envelope* or subset of states that might be relevant to the planning problem at hand (e.g., given particular initial and goal states) is constructed, and an optimal policy is computed for this restricted space in an anytime fashion. Clearly, optimality is sacrificed since important states might lie outside the envelope, as is generality, for the policy makes no mention of these ignored states. In (Dean et al. 1993b) it is suggested that domain-specific heuristics will aid in initial envelope selection and envelope alteration.

We propose an alternative method for dealing with Markov decision models in a real-time environment. We suggest that MDPs be explicitly viewed as search problems. Real-time constraints can be incorporated by restricting the search horizon. This is the basic idea behind, for example, Korf's real-time heuristic search algorithm (1990). In stochastic domains there is another important reason for interleaving execution into the planning process, namely, to restrict the search space to the *actual* outcomes of probabilistic actions. In particular, once a certain action is deemed best (for a given state) it should be executed and its outcome observed. Subsequent search for the best next action can proceed from the actual outcome, ignoring other unrealized outcomes of that action. To further improve real-time performance, we can cache the best action for a state once it has been computed and use the cached value if the agent returns to that state.

In general, a fixed-depth search will tend to be greedy, choosing actions that provide immediate reward at the expense of long-term gain. To alleviate this problem we assume a heuristic function that estimates the *value* of each state, accounting for future states that might be reached in addition to that state's immediate reward. This prevents (to some extent) the problem of globally suboptimal choices due to finite horizon effects. Knowledge of certain properties of the heuristic function allow the search tree to be pruned. We describe one method of constructing heuristic functions that allows this pruning information to be easily determined. This construction also produces default actions for each state, in essence, generating a reactive policy. Our search procedure can be viewed as using deliberation to refine the reactive strategy.

The next section presents the MDP model in detail and also looks at a more natural representation for describing worlds and actions. Section 3 describes the algorithm, the interleaving of search and execution, and possible pruning methods. Section 4 discusses ways of constructing the heuristic evaluation functions, and Section 5 examines the computational cost of the algorithm, and provides some experimental results.

## 2 The Decision Model

Let $\mathcal{S}$ be a finite set of world states. In many domains the states will be the models (or worlds) associated with some logical language, so $|\mathcal{S}|$ will be exponential in the number of atoms generating this language. Let $\mathcal{A}$ be a finite set of actions available to an agent. An action takes the agent from one world to another, but the result of an action is known only with some probability. An action may then be viewed as a mapping from $\mathcal{S}$ into probability distributions over $\mathcal{S}$. We write $Pr(s_1, a, s_2)$ to denote the probability that $s_2$ is reached given that action $a$ is performed in state $s_1$ (embodying the usual Markov assumption). We assume that an agent, once it has performed an action, can observe the resulting state; hence the process is *completely observable*. Uncertainty in this model results only from the outcomes of actions being probabilistic, not from uncertainty about the state of the world. We assume a real-valued *reward function* $R$, with $R(s)$ denoting the (immediate) utility of being in state $s$. For our purposes an MDP consists of $\mathcal{S}$, $\mathcal{A}$, $R$ and the set of transition distributions $\{Pr(\cdot, a, \cdot) : a \in \mathcal{A}\}$.

A control *policy* $\pi$ is a function $\pi : \mathcal{S} \to \mathcal{A}$. If this policy is adopted, $\pi(s)$ is the action an agent will perform whenever it finds itself in state $s$. Given an MDP, an agent ought to adopt an optimal policy that maximizes the expected rewards accumulated as it performs the specified actions. We concentrate here on *discounted infinite horizon* problems: the value of a reward is discounted by some factor $\beta$ $(0 < \beta < 1)$ at each step in the future; and we want to maximize the expected accumulated discounted rewards over an infinite time period. Intuitively, a DTP problem can be viewed as finding a good (or optimal) policy.

The expected *value* of a fixed policy $\pi$ at any given state $s$ is specified by

$$V_\pi(s) = R(s) + \beta \sum_{t \in \mathcal{S}} Pr(s, \pi(s), t) \cdot V_\pi(t)$$

Since the factors $V_\pi(s)$ are mutually dependent, the value of $\pi$ at any initial state $s$ can be computed by solving this system of linear equations. A policy $\pi$ is *optimal* if $V_\pi(s) \geq V_{\pi'}(s)$ for all $s \in \mathcal{S}$ and policies $\pi'$.

Although we represent actions as sets of stochastic transitions from state to state, we expect that domains and actions will usually be specified in a more traditional form for planning purposes. Figure 1 shows a stochastic variation of STRIPS rules (Kushmerick, Hanks and Weld 1993) for a domain in which the robot must deliver coffee to the user. An effect $E$ is a set of literals. If we apply $E$ to some state $s$, the resulting state satisfies all the literals in $E$ and agrees with $s$ for all other literals. The *probabilistic effect* of an action is a finite set $E_1, ... E_n$ of effects, with associated probabilities $p_1, ..., p_n$ where $\sum p_i = 1$.

Since actions may have different results in different contexts, we associate with each action a finite set $D_1, ..., D_n$ of mutually exclusive and exhaustive sentences called *discriminants*, with probabilistic effects $EL_1, ..., EL_n$. If the action is performed in a state $s$ satisfying $D_i$, then a random effect from $EL_i$ is applied to $s$. For example, in Figure 1, if the agent carries out the *GetUmbrella* action in a state where *Office* is true, then with probability 0.9 *Umbrella* will be true, and every other proposition will remain unchanged, and with probability 0.1 there will be no change of state. For convenience, we may also write actions as sets of *action aspects* as illustrated for the *Move* action in Figure 1 (Boutilier and Dearden 1994). The action has two descriptions which represent two independent sets of discriminants and the cross product of the aspects is used to determine the actual effects. For example, if *Rain* and *Office* are true, and a *Move* action is performed then with probability 0.81 $\neg$*Office*, *Wet* will result, and so on. Action aspects allow independence to be represented explicitly, in a similar manner to causal networks. The representation of domains in terms of propositions also provides a natural way of expressing rewards. Figure 2 shows a representation of rewards for this domain. The reward for any given state depends only on the values of the propositions *HasUserCoffee* and *Wet* in that state.

This framework is flexible enough to allow a wide variety of different reward functions. One important situation is that in which there is some set $\mathcal{S}_g \subseteq \mathcal{S}$ of goal states, and the agent tries to reach a goal state in as few moves as possible.[1] Since we are interleaving plan construction and plan execution, the time required to plan is significant when measuring success; but as a first approximation we can represent this type of situation with the reward function (Dean et al. 1993b): $R(s) = 0$ if $s \in S_g$ and $R(s) = -1$ otherwise.

| Action | Discriminant | Effect | Prob |
|---|---|---|---|
| Move | Office | ¬Office | 0.9 |
|  |  |  | 0.1 |
|  | ¬Office | Office | 0.9 |
|  |  |  | 0.1 |
| Move | Rain,¬Umb | Wet | 0.9 |
|  |  |  | 0.1 |
| BuyCoffee | ¬Office | HRC | 0.8 |
|  |  |  | 0.2 |
|  | Office |  | 1.0 |
| GetUmb. | Office | Umbrella | 0.9 |
|  |  |  | 0.1 |
| DelCoffee | Office,HRC | HUC,¬HRC | 0.8 |
|  |  | ¬HRC | 0.1 |
|  |  |  | 0.1 |
|  | ¬Office,HRC | ¬HRC | 0.8 |
|  |  |  | 0.2 |
|  | ¬HRC |  | 1.0 |

Figure 1: An example domain presented as STRIPS-style action descriptions. Note that HUC and HRC are HasUserCoffee and HasRobotCoffee respectively.

| Conditions | Reward | Conditions | Reward |
|---|---|---|---|
| HUC, ¬Wet | 1.0 | HUC, Wet | 0.8 |
| ¬HUC, ¬Wet | 0.2 | ¬HUC, Wet | 0.0 |

Figure 2: An example of a reward function for the coffee delivering robot domain.

## 3 The Algorithm

Our algorithm for integrating planning and execution proceeds by searching for a best action, executing that action, observing the result of this execution, and iterating. The underlying search algorithm constructs a partial decision tree to determine the best action for the current state (the root of this tree). We assume the existence of a heuristic function that estimates the value of each state (such heuristics are described in Section 4). The search tree may be pruned if certain properties of the heuristic function are known. This search can be terminated when the tree has been expanded to some specified depth, when real-time pressures are brought to bear, or when the best action is known (e.g., due to complete pruning, or because the best action has been cached for this state).

Once the search algorithm selects a best action for the current state, the action is executed and the resulting state is observed. By observing the new state, we establish which of the possible action outcomes actually occurred. Without this information, the search for the best *next* action would be forced to account for every possible outcome of the previous action. By interleaving execution and observation with search, we need only search from the *actual* resulting state.

In skeletal form, the algorithm is as follows. We denote by $s$ the current state, and by $A^*(t)$ the best action for state $t$.[2]

1. If state $s$ has not been previously visited, build a *partial decision tree* of all possible actions and their outcomes beginning at state $s$, using some criteria to decide when to stop expanding the leaves of the tree. Using the partial tree and the heuristic function, calculate the best action $A^*(s) \in \mathcal{A}$ to perform in state $s$. (This value may be cached in case $s$ is revisited.)

2. Execute $A^*(s)$.

3. Observe the actual outcome of $A^*(s)$. Update $s$ to be this observed state (the state is known with certainty, given the assumption of complete observability).

4. Repeat.

---

[1] If a "final" state stops the process, we may use self-absorbing states or include a null action which does nothing.

[2] Initially $A^*(t)$ might be undefined for all $t$. However, if the heuristic function provides default reactions (see Section 4), it is useful to think of these as the best actions determined by a depth 0 search.

The point at which the algorithm stops depends on the characteristics of the domain. For example, if there are goal states, and the agent's task is to reach one, planning may continue until a goal state is reached, while in process-oriented domains, the algorithm continues indefinitely. In our experiments, we have typically run the algorithm for a constant number of steps. Note that by caching the best action for each state, the agent slowly constructs a policy for all the states it is likely to reach. The use of caching here is similar to that of Learning RTA* search (Korf 1990).

### 3.1 Action Selection

Here we discuss step one of the high-level algorithm given above. To select the best action for a given state, the agent needs to estimate the value of performing each action. In order to do this, it builds a partial decision tree of actions and resulting states, and uses the tree to approximate the expected utility of each action. This search technique is related to the *-minimax algorithm of Ballard (1983). As we shall see in Section 3.2, there are similarities in the way we can prune the search tree as well. Figure 3 shows a partial tree of actions two levels deep. From the initial state $s$, if we perform action $A$, we reach state $t$ with probability 0.8, and state $u$ with probability 0.2. The agent expands these states with a second action to produce the set of second states. To determine the action to perform in a given state, the agent estimates the expected utility of each action. If $s$ and $t$ are states, $\beta$ is the factor by which the reward for future states is discounted, and $\mathcal{V}(t)$ is the heuristic function at state $t$, the *estimated* expected utility of action $A_i$ is:

$$U(A_i|s) = \sum_{t \in S} Pr(s, A_i, t) V(t)$$

$$V(s) = \begin{cases} \mathcal{V}(s) & \text{if } s \text{ is a leaf node} \\ R(s) + \beta \cdot \max\{U(A_j|s) : A_j \in \mathcal{A}\} & \text{otherwise} \end{cases}$$

Figure 3 illustrates the process with a discounting factor of 0.9. The utility of performing action $A$ if the world were in state $t$ is the weighted sum of the values of being in states $x$ and $y$, which is 2.1. Since the utility of action $B$ is 0.3, we select action $A$ as the best (given our current information) for state $t$, and make $V(t) = R(t) + \beta U(A|t) = 2.39$. The utility of action $A$ in state $s$ is $Pr(s, A, t)V(t) + Pr(s, A, u)V(u)$, giving $U(A|s) = 2.23$. This is lower than $U(B|s)$, so we select $B$ as the best action for state $s$, record the fact that $A^*(s)$ is $B$, and execute $B$. By observing the world, the agent now knows whether state $v$ or $w$ is the new state, and can build on its previous tree, expanding the appropriate branch to two levels and determining the best action for the new state. Notice that if (say) $v$ results from action $B$, the tree rooted at state $w$ can safely be ignored — the unrealized possibility can have no further impact on updated expected utility (unless $w$ is revisited via some path).

If the agent finds itself in a state visited earlier, it may use the previously calculated and cached best action $A^*(s)$. This avoids the recalculation of visited states, and considerably speeds planning if the same or related problems must be solved multiple times, or if actions naturally lead to "cycles" of states. Eventually, $A^*(s)$ could contain a policy for every reachable state in $S$, removing the need for further computation.

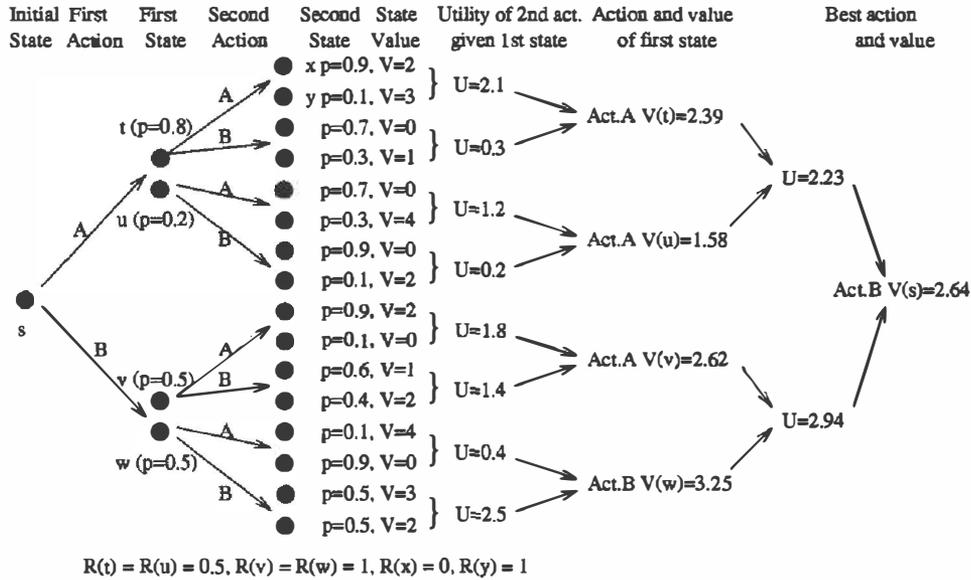

Figure 3: An example of a two-level search for the best action from state $s$.

In Figure 3, the tree is expanded to depth two. The depth can obviously vary depending on the available time for computation. The deeper the tree is expanded, the more accurate the estimates of the utilities of each action tend to be, and hence the more confidence we should have that the action selected approaches optimality. Indeed, it is quite natural to view the search process as a directed form of value iteration. The heuristic serves as the initial value vector, and a step of the search corresponds to a partial update of the vector. Convergence results for value iteration can be adapted to this setting.

If there are $m$ actions, and the number of states that could result from executing an action is on average $b$, then a tree of depth one will require $O(mb)$ steps, two levels will require $O(m^2b^2)$, and so on. The potentially improved performance of a deeper search has to be weighed against the time required to perform the search (Russell and Wefald 1991). Rather than expand to a constant depth, the agent could instead keep expanding the tree until the probability of reaching the state being considered drops below a certain threshold. This approach may work well in domains where there are extreme probabilities or utilities.

### 3.2 Techniques for Limiting the Search

As it stands, the search algorithm performs in a very similar way to minimax search. Determining the value of a state is analogous to the MAX step in minimax, while calculating the value of an action can be thought of as an AVERAGE step, which replaces the MIN step (see also (Ballard 1983)). When the search tree is constructed, we can use techniques similar to those of Alpha-Beta search to prune the tree and reduce the number of states that must be expanded. There are two applicable pruning techniques. To make our description clearer, we will treat a single ply of search as consisting of two steps, $MAX$ in which all the possible actions from a state are compared, and $AVERAGE$, where the outcomes of a particular action are combined. Two sorts of cuts can be made in the search tree. If we know bounds on the maximum and/or the minimum values of the heuristic function, *utility cuts* (much like $\alpha$ and $\beta$ cuts in minimax search) can be used. If the heuristic function is reasonable, the maximum and minimum values for any state can be bounded easily using knowledge of the underlying decision process. In particular, with maximum and minimum immediate rewards of $R^+$ and $R^-$, the maximum and minimum expected values for any state are bounded by $\frac{1}{1-\beta} \cdot R^+$ and $\frac{1}{1-\beta} \cdot R^-$, respectively. If we have bounds on the error associated with the heuristic function, *expectation cuts* may be applied. These are illustrated with examples.

**Utility Pruning** We can prune the search at an AVERAGE step if we know that no matter what the value of the remaining outcomes of this action, we can never exceed the utility of some other action at the preceding MAX step. For example, consider the search tree in Figure 4(a). We assume that the maximum value the heuristic function can take is 10. When evaluating action $b$, since we know that the value of the subtree rooted at $T$ is 5, and the best that the subtrees below $U$ and $V$ could be is $0.1 \times 10 + 0.2 \times 10 = 3$, the total cannot be larger than $3.5 + 1 + 2 = 6.5$ so neither the tree below $U$ nor that below $V$ is worth expanding. This type of pruning requires that we know in advance the maximum value of the heuristic function. The minimum value can be used in a more restricted fashion. It performs especially well when nodes are ordered for expansion according to their probability of actually occurring.

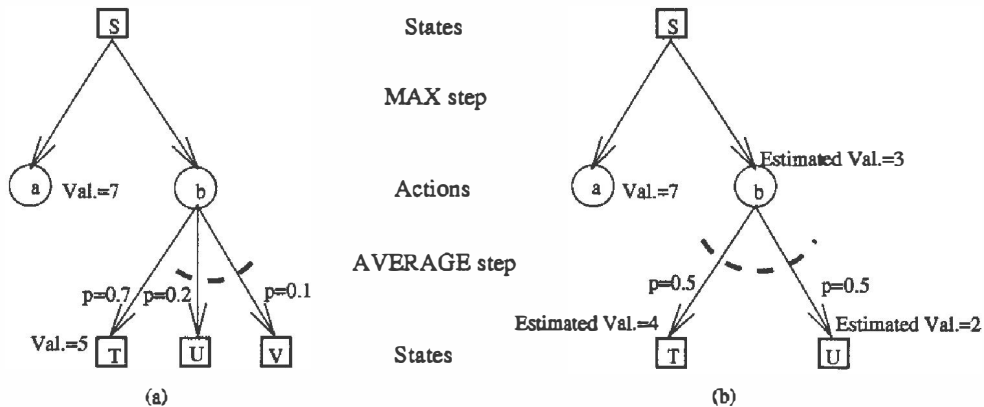

Figure 4: Two kinds of pruning where $\mathcal{V}(s) \leq 10$ and is accurate to $\pm 1$. In (a), utility pruning, the trees at $U$ and $V$ need not be searched, while in (b), expectation pruning, the trees below $T$ and $U$ are ignored, although the states themselves are evaluated.

**Expectation Pruning** For this type of pruning, we need to know the maximum error associated with the heuristic function (see (Boutilier and Dearden 1994) for a way of estimating this value). If we are at a maximizing step and, even taking into account the error in the heuristic function, the action we are investigating cannot be as good as some other action, then we do not need to expand this action further. For example, consider Figure 4(b), where we assume that $\mathcal{V}(S)$ is within $\pm 1$ of its true (optimal) value. We have determined that $U(a|S) = 7$, therefore any potentially better action must have a value greater than 6. Since $p(S, a, T)\mathcal{V}(T) + p(S, a, U)\mathcal{V}(U) \leq 4$, even if $b$ is as good as possible (given these estimates), it cannot achieve this threshold, so there is no need to search further below $T$ and $U$.

Expectation pruning requires a modification of the search algorithm to check all outcomes of an action to see if the weighted average of their estimated values is sufficient to justify continued node expansion. This means that the heuristic value of sibling nodes must be checked before expanding a given node, and a tight bound on $\mathcal{V}$ must be found. However, the method of generating heuristic functions described in (Boutilier and Dearden 1994) (see the next section for a brief discussion) produces just such bounds. Expectation pruning is closely related to what Korf (1990) calls alpha-pruning. The difference is that while Korf relies on a property of the heuristic that it is always increasing, we rely on an estimate of the actual error in the heuristic.

## 4 Generating Heuristic Functions

We have assumed the existence of a heuristic function above. We now briefly describe some possible methods for generating these heuristics. The problem is to build a heuristic function which estimates the value of each state as accurately as possible with a minimum of computation. In some cases such a heuristic may already be available. Here we will sketch an approach for domains with certain characteristics, and suggest ideas for other domains.

### 4.1 Abstraction by Ignoring Propositions

In certain domains, actions might be represented as STRIPS-like rules as in Figure 1, and the reward function specified in terms of certain propositions. If this is the case we can build an abstract representation of the state space by constructing a set $\mathcal{R}$ of *relevant* propositions, and using it to construct abstract states each corresponding to all the states which agree on the values of the propositions in $\mathcal{R}$. A complete description of our approach, along with theoretical and experimental results, can be found in (Boutilier and Dearden 1994). However we will broadly describe the technique here.

To construct $\mathcal{R}$, we first construct a set of *immediately relevant* propositions $\mathcal{IR}$. These are propositions that have significant effect on the reward function. For example, in Figure 2, both *HasUserCoffee* and *Wet* have an effect on the reward function; but to produce a small abstract state space, $\mathcal{IR}$ might include only *HasUserCoffee*, since this is the proposition which has the greatest effect on the reward function.

$\mathcal{R}$ will include all the propositions in $\mathcal{IR}$, but also any propositions that appear in the discriminant of an action which allows us to change the truth value of some proposition in $\mathcal{R}$. Formally, $\mathcal{R}$ is the smallest set such that: 1) $\mathcal{IR} \subseteq \mathcal{R}$; and 2) if $P \in \mathcal{R}$ occurs in an effect list of some action, then all propositions in the corresponding discriminant are in $\mathcal{R}$.

$\mathcal{R}$ induces a partition of the state space into sets of states, or *clusters* which agree on the truth values of propositions in $\mathcal{R}$. Furthermore, the actions from the original 'concrete' state space apply directly to these clusters. This is due to the fact that each action either maps all the states in a cluster to the same new cluster, or changes the state, but leaves the cluster unchanged. These two facts allow us to perform policy iteration on the abstract state space. The algorithm is:

1. Construct the set of relevant propositions $\mathcal{R}$. The actions are left unchanged, but effects on propositions not in $\mathcal{R}$ are ignored.
2. Use $\mathcal{R}$ to partition the state space into clusters.
3. Use the policy iteration algorithm to generate an abstract policy for the abstract state space. For details of this algorithm see (Howard 1971; Dean et al. 1993b).

By altering the number of reward-changing propositions in $\mathcal{R}$, we can vary its size, and hence the granularity and accuracy of the abstract policy. This allows us to investigate the tradeoff between time spent building the abstract policy and its degree of optimality. The policy iteration algorithm also computes the value of each cluster in the abstract space. This value can be used as a heuristic estimate of the value of the cluster's constituent states. One advantage of this approach is that it allows us to accurately determine bounds on the difference between the heuristic value for any state, and its value according to an optimal policy — see (Boutilier and Dearden 1994) for details. As shown above, this fact is very useful for pruning the search tree. A second advantage of this method for generating heuristic values is that it provides default reactions for each state.

### 4.2 Other Approaches

The algorithm described above for building the heuristic function is certainly not appropriate in all domains. Certain domains are more naturally represented by other means (navigation is one example). In some cases abstractions of actions and states may already be available (Tenenberg 1991).

For robot navigation tasks, an obvious method for clustering states is based on geographic features. Nearby locations can be clustered together into states that represent regions of the map, but providing actions that operate on these regions is more complex. One approach is to assume some probability distribution over locations in each region, and build abstract actions as weighted averages over all locations in the region of the corresponding concrete action. The difficulty with this approach is that it is computationally expensive, requiring that every action in every state be accounted for when constructing the abstract actions.

If abstract actions (possibly macro-operators (Fikes and Nilsson 1971)) are already available, we need to find clusters to which the actions apply. In many cases this may be easy as the abstract actions may treat many states in exactly the same way, hence generating a clustering scheme. In other domains, a similar weighted average approach may be needed.

## 5 Theoretical and Experimental Results

We are currently exploring, both theoretically and experimentally, the tradeoffs involved in the interleaving of planning and execution in this framework. We can measure the complexity of the algorithm as presented. Let $m = |\mathcal{A}|$ be the number of actions. We will assume that when constructing the search tree for a state, we explore to depth $d$, and that the branching factor for each action (the maximum number of outcomes for the action in any given state) is at most $b$.[3]

The cost of calculating the best action for a single state is $m^d b^d$. The cost per state is slightly less than this since we can reuse our calculations, but the overall complexity is $O(m^d)$. The actual size of the state space has no effect on the algorithm; rather it is the number of states visited in the execution of the plan that affects the cost. This is clearly domain dependent, but in most domains should be considerably lower than the total number of states. Most importantly, the complexity of the algorithm is constant and execution time (per action) can be bounded for a fixed branching and search depth. By interleaving execution with search, the search space can be drastically reduced. When planning for a sequence of $n$ actions the execution algorithm is linear in $n$ (with respect to the factor $m^d b^d$); a straightforward search without execution for the same number of actions is $O(b^n)$.

Experiments in a number of different domains provide an indication that this framework may be quite valuable. To generate the results discussed in this section, we used a domain based on the one described in Figures 1 and 2 but with another item (snack) that the robot must deliver, and a robot that only carries one thing at a time. We constructed the heuristic function using the procedure described in Section 4, with *HasUserCoffee* as the only immediately relevant proposition, ignoring the proposition *HasUserSnack*; thus, $\mathcal{R} = \{$HasUserCoffee, Office, HasRobotCoffee, HasRobotSnack$\}$.

The domain contains 256 states and six actions. All timing results were produced on a Sun SPARCstation 1. Computing an optimal policy by policy iteration required 130.86 seconds, while computing a sixteen state abstract policy (again using policy iteration) for the heuristic function required 0.22 seconds. Figure 5a. shows the time required to search for the best action in a single state, and the expected reward from the state (¬HasRobotCoffee, ¬HasUserCoffee, ¬HasRobotSnack, ¬HasUserSnack, Rain, ¬Umbrella, Office, ¬Wet)[4]. Time to search grows exponentially with search depth, while the reward steadily improves until it reaches its maximum value. As the graph shows, a close to optimal policy for this particular start state was found with depth 4 search. As searching to depth 4 required less than 5 seconds per state, this is a considerable saving over policy iteration. As the domain grows in size, deeper search may be necessary to produce close to optimal behavior, but the time required for policy iteration typically grows at a faster rate, so the cost of deeper search is justified.

---

[3] We ignore preconditions for actions here, assuming that an action can be "attempted" in any circumstance. However, preconditions may play a useful role by capturing user-supplied heuristics that filter out actions in situations in which they *ought not* (rather than *cannot*) be attempted. This will effectively reduce the branching factor of the search tree.

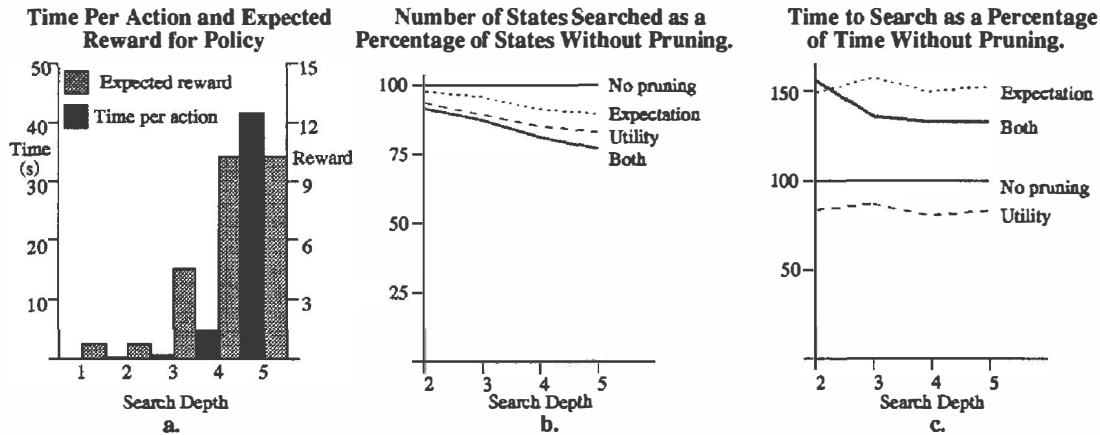

Figure 5: a.: Time to search for the best action and expected reward for one particular state. b., c.: Value of pruning search over standard search. Both as a function of search depth. The domain contains 256 states and six actions.

Figure 5b. and c. show the effects of pruning on the performance of the algorithm. 5b. shows the percentage of states which are searched as search depth increases. *Utility pruning* performs better than *expectation pruning*, removing about 20 percent of states when searching to depth five. 5c. shows the time required to search. *Utility pruning* again performs well, with a 15 to 20 percent saving in execution time. Although the complexity of performing expectation pruning results in a slower performance than no pruning at all, in deep search, it may well be worthwhile to perform expectation pruning close to the root of the tree where its effect will be the greatest. Value of computation calculations (1991) might be used to determine the point at which to stop expectation pruning.

Figure 6 shows a variety of statistics about the policy induced by various depths of search. There is a steady improvement in the quality of the policy as search depth increases, with an almost optimal policy being discovered at depth 5. Since the value of a state can range from −15 to 20 in this domain, the errors made in even relatively shallow search are quite small. The table also suggests that searching to depth $n$ is at least as good as searching to depth $n - 1$ (although it may not always be better). In none of the domains we have tested has searching deeper produced a worse policy, although this may not be the case in general (see (Pearl 1984) for a proof of this for minimax search).

| Search Depth | 1 | 2 | 3 | 4 | 5 |
|---|---|---|---|---|---|
| No. of errors | 137 | 137 | 132 | 22 | 8 |
| Total Error | 714 | 589 | 549 | 35.7 | 3.4 |
| Max. Error | 12.5 | 9.4 | 8.2 | 7.3 | 0.5 |
| Average Error | 2.8 | 2.3 | 2.1 | 0.1 | 0.01 |

Figure 6: A comparison of policy quality. The policies are compared with the optimal policy for the domain. Here errors are states where the policy constructed by searching and the optimal policy disagree on the value of the state.

Figure 7 shows the time required for searching with and without execution, and with and without caching. The search-execution model we have investigated performs better than straight search, although, especially for deep search, the difference is fairly small. This is due to the small size of the domain and the effects of caching, which allow the search without execution algorithm to only search below each state once.

## 6   Conclusions

We have proposed a framework for planning in stochastic domains. Further experimental work needs to be done to demonstrate the utility of this model. In particular, further comparison to exact methods like policy iteration and heuristic methods like the envelope approach of Dean et al. (1993b) would be useful. We intend to use the framework to explore a number of tradeoffs (e.g., as in (Russell and Wefald 1991)). In particular, we will look at the advantages of a deeper search tree, and balance this with the cost of building such a tree, and at the tradeoff between computation time and improved results when building the heuristic function. To illustrate these ideas we observe that if the depth of the search tree is 0, this corresponds to a reactive system where the best action for each state is obtained from the abstract policy. If each cluster for the abstract policy contains a single state, we have optimal policy planning. The usefulness of these tradeoffs will vary when planning in different domains.

---

[4]This is the state that requires the longest sequence of actions to reach a state with maximal utility; i.e., the state requiring the "longest optimal plan."

| Depth | Search | Search-execution | No cache | Search, no cache |
|---|---|---|---|---|
| 1 | 5.19 | 0.01 | 0.02 | 26.7 |
| 2 | 5.41 | 0.04 | 0.06 | 281 |
| 3 | 7.14 | 0.42 | 0.51 | 2780 |
| 4 | 15.4 | 4.48 | 5.68 | - |
| 5 | 102 | 55.4 | 56.9 | - |
| 6 | 272 | 219 | 230 | - |

Figure 7: Time comparisons in seconds for search without execution, search with execution, search with execution but no caching, and search without execution or caching. All searches were performed as if 10 actions were selected.

Some of the characteristics of domains that will affect our choices are:

- *Time:* for time-critical domains it may be better to limit time spent deliberating (perhaps adopting a reactive strategy based on the heuristic function). A more detailed heuristic function and a smaller search tree may be appropriate.

- *Continuity:* if actions have similar effects in large classes of states and most of the goal states are fairly similar, we can use a less detailed heuristic function (more abstract policy).

- *Fan-out:* if there are relatively few actions, and each action has a small number of outcomes, we can afford to increase the depth of the search tree.

- *Plausible goals:* if goal states are hard to reach, a deeper search tree and a more detailed heuristic function may be necessary.

- *Extreme probabilities:* with extreme probabilities it may be worth only expanding the tree for the most probable outcomes of each action. This seems to bear some relationship to the envelope reconstruction phase of the recurrent deliberation model of Dean et al. (1993a).

In the future we hope to continue our experimental investigation of the algorithm to look at the effectiveness of variable-depth search, and the possibility of improving the heuristic function by recording newly computed values of states, rather than best actions. We also hope to investigate the performance of this approach in other types of domains, including high-level robot navigation, and scheduling problems, and to further investigate the theoretical properties of the algorithm, especially through analysis of the value of deeper searching in producing better plans (Pearl 1984). Our model can be extended by relaxing some of the assumptions incorporated into the decision-model. Semi-markov processes as well as partially observable processes will require interesting modifications of our model. Finally, we must investigate the degree to which the restricted envelope approach may be meshed with our model.

## Acknowledgments

Discussions with Moisés Goldszmidt have considerably influenced our view and use of abstraction for MDPs. This research was supported by NSERC Research Grant OGP0121843 and a UBC University Graduate Fellowship.